\documentclass[a4paper, 11pt, twoside]{article}

\usepackage{graphicx, psfrag, epsfig, amssymb, amsthm, float, booktabs, cite, url, datetime, parskip, fancyhdr, hyperref}
\usepackage[ruled, noline, linesnumbered]{algorithm2e}
\usepackage[margin=1in]{geometry}
\usepackage[labelfont=bf]{caption}

\SetKwProg{Fn}{Function}{}{end}\SetKwFunction{GetCoordinates}{GetCoordinates}

\AtBeginDocument{%
  \providecommand\BibTeX{{%
    \normalfont B\kern-0.5em{\scshape i\kern-0.25em b}\kern-0.8em\TeX}}}

\theoremstyle{definition}
\newtheorem{example}{Example}[section]

\theoremstyle{definition}
\newtheorem{definition}{Definition}[section]

\newdateformat{monthyeardate}{%
  \monthname[\THEMONTH] \THEYEAR}

\newcommand\correspondingauthor{\thanks{Corresponding author.}}
\begin{document}
\monthyeardate
\title{Coordinate Matrix Machine: A Human-level Concept Learning to Classify Very Similar Documents}

\author{Amin Sadri and M Maruf Hossain\correspondingauthor \\
{\small \texttt{\{amin12sadri, maruf.hossain.phd\}@gmail.com}}
}

\date{}

\maketitle
\thispagestyle{empty}

\begin{abstract}
\noindent Human-level concept learning argues that humans typically learn new concepts from a single example, whereas machine learning algorithms typically require hundreds of samples to learn a single concept. Our brain subconsciously identifies important features and learns more effectively.
\vspace*{6pt}

\noindent \textbf{Contribution:} In this paper, we introduce Coordinate Matrix Machine (CM$^{2}$), a structure-aware document intelligence model designed to bridge the gap between semantic content and spatial layout. Unlike traditional NLP approaches that prioritize sequential text, CM$^{2}$ is particularly effective for low-data, layout-sensitive tasks, enabling high-precision performance when structural geometry is paramount. While modern ``Red AI’’ trends rely on massive pre-training and energy-intensive GPU infrastructure, CM$^2$ is designed as a \textbf{Green AI solution}. It achieves human-level concept learning by identifying only the structural ``important features’’ a human would consider, allowing it to classify very similar documents using only one sample per class.
\vspace*{6pt}

\noindent \textbf{Advantage:} 
Our algorithm outperforms traditional vectorizers and complex deep learning models that require larger datasets and significant compute. By focusing on structural coordinates rather than exhaustive semantic vectors, CM$^2$ offers:
\begin{enumerate}
\item High accuracy with minimal data (one-shot learning)
\item Geometric and structural intelligence
\item Green AI and environmental sustainability
\item Optimized for CPU-only environments
\item Inherent explainability (glass-box model)
\item Faster computation and low latency
\item Robustness against unbalanced classes
\item Economic viability
\item Generic, expandable, and extendable
\end{enumerate}
\vspace*{6pt}

\noindent \textbf{Source Code}: The source code for the Python implementation of CM$^2$ is available on \href{https://github.com/maruf42/Coordinate-Matrix-Machine/}{GitHub Repository}.
\vspace*{6pt}

\end{abstract}

\section{Introduction}
Human-level concept learning is a relatively less explored area of research. The goal is to provide a solution to a problem that is easy for humans to do, yet still challenging for machines despite their computational power. People can learn new concepts from only one or a very few samples, whereas machine learning methods require many examples to discover correlations and understand features. The reason is that people subconsciously select the important features and then generalize to other distinct areas~\cite{lake2015,mnih2015}.

In this paper, we deploy a human-level concept learning approach for document classification. Document classification is a common task in machine learning and Natural Language Processing (NLP), in which a document is assigned to one or more classes. Current approaches rely heavily on the document context to classify documents~\cite{jones1972,power2010}. Identifying the relevant topic is the primary motivator for such document classification~\cite{hassan2017,kotenko2015,lin2006}. These models assume we have access to ample labelled data and that the document context is sufficiently informative to distinguish between classes. Some other methods use document image information, thereby exacerbating the labeling process. Since images are high-dimensional data, labeling so many training samples is not feasible in most real-world problems~\cite{krizhevsky2012}. While most approaches treat text in isolation, our solution rests on the premise that the meaning of a structured document is derived from both the text and its coordinates. This dual modality allows the model to leverage spatial grounding that purely text-based models overlook.

A practical scenario is receiving bank statements from other financial institutions along with loan applications; the task is to identify the statement templates so that existing functions can be applied to extract information from them. In practice, when we ask someone to classify these statements, we do not need to show them hundreds of samples per class. A single sample suffices to represent a class. This fact indicates that a single sample is sufficient for classification if the algorithm is well-designed. This is our motivation for developing our algorithm. 

\textit{Our aim is to develop a coordinate matrix-based framework (Coordinate Matrix Machine, CM$^2$) as a purpose-built small model that achieves human-level concept learning without the prohibitive costs of Large Language Models (LLMs). Unlike `Red AI' trends that rely on massive pre-training and energy-intensive GPU infrastructure, CM$^2$ is designed as a Green AI solution. It prioritizes computational efficiency and environmental sustainability by mimicking human structural identification rather than processing exhaustive semantic vectors. This enables a high-precision, explainable alternative that is both economically viable and runs on standard CPU hardware.}

\subsection{Challenges}
When there is substantial variation in structured documents whose structure must be classified, machine learning techniques often perform poorly. We have identified the following challenges when dealing with bank statements in particular:
\begin{itemize}
\item A large number of documents need to be labelled to train a model with sufficient accuracy, which is both time and resource-consuming.
\item There are way too many class labels at the template level. For the five Australian banks included in this experiment, 53 templates (classes) were available. There are not enough representative samples for each class.
\item All bank statements are very similar in context, as they include many transaction descriptions, dates, and amounts.
\item Majority of the words in bank statements are personal and highly contextual, such as account holder’s name, address, or transaction items, which produces lots of noise when training a classification model.
\item The words which are not noise are very similar across a range of bank statements, e.g., `Name’, `Account’, `Balance’, `Date’.
\end{itemize}

Due to these challenges, traditional machine learning models and more recent deep learning models perform poorly. 
\begin{example}
When classifying statements using the Term Frequency Vectorizer with Logistic Regression, the model with the best parameters achieved an $F$-measure of 79\% on the templates over 53 templates in an 80-20 training-test split.
\end{example}

This lower performance can discourage users from using simplistic models and incline them to use more complex models like Convolutional Neural Network (CNN) or Recurrent Neural Network (RNN) \cite{banerjee2019,kumar2019}, which are by default not explainable, and rely on additional techniques~\cite{ribeiro2016} to explain, especially to the governing body for compliance purposes.
\begin{example}
After spending hours building a CNN or Long Short-Term Memory (LSTM) RNN architecture, we found that the CNN achieved an $F$-measure of 93\%, whereas the LSTM failed to classify 53 templates.
\end{example}

\subsection{Our Contribution}
In this paper, we apply a human-level concept learning~\cite{lake2015} to address this problem. We choose this approach because, as humans, we do not need hundreds of training samples to learn a single class, and we can classify documents even when the content is similar, sometimes even without reading the texts. Only one sample per template is sufficient, and we should consider not only the content but also the locations of specific words.

The work closest to ours in approach is that of Lake et al.~\cite{lake2015}. They have argued that in most cases, people can learn a new concept from just one or a handful of examples. At the same time, typical machine learning algorithms require hundreds of examples to perform similarly. They aim to address a different problem: recognising handwritten characters from image data. In this paper, we position CM$^{2}$ as a specialized solution optimized for entity extraction (not covered in this paper) and document classification in complex structural domains. By treating the document as a coordinate matrix rather than a linear sequence, CM$^{2}$ provides the ‘structural intelligence’ necessary for high-stakes industrial processing. Although the issues differ, both works use human-level concept learning by examining how humans solve the problem. Both approaches only use one sample per class and learn from that sample.

Given documents of similar structure, as humans, we examine the document structure and/or the positions of specific keywords (e.g., Account Name, Date, Bank Name). Similarly, in our approach, we propose a hybrid lazy learning algorithm, the Coordinate Matrix Machine (CM$^2$), that constructs an input matrix (in contrast to traditional vector-based input strategies) containing the coordinates of the keywords and uses this matrix to classify documents.

The rationale behind the technique is as follows:
\begin{enumerate}
\item With structured documents, the position carries more importance than the occurrence or the order of the occurrence for the terms. Furthermore, by augmenting the subject-matter experts’ understanding, we can avoid building a large corpus, thereby reducing noise and improving classification performance.
\item This technique would allow us to avoid labeling a large number of documents, which is often required to train machine learning and deep learning models.
\end{enumerate}

\subsection{Advantages}
The advantages of our algorithm are:
\begin{enumerate}
\item \textbf{High Accuracy with Minimal Data (One-Shot Learning):} The algorithm achieves human-level concept learning, successfully classifying documents using only a single sample per class. It outperforms traditional machine learning and advanced deep learning models that typically require hundreds or thousands of labeled examples to achieve comparable accuracy. This makes the framework particularly effective for low-data, layout-sensitive tasks where obtaining a large, diverse training corpus is infeasible.
\item \textbf{Geometric and Structural Intelligence:} Unlike text-only transformer-based models (like BERT or GPT) that process text as a linear sequence of tokens, CM$^2$ utilizes a coordinate matrix to capture the physical geometry of the document. This makes it significantly more effective for formal documents (bank statements, invoices) where the spatial position of elements is more informative than semantic flow. As a structure-aware document intelligence model, CM$^{2}$ ensures that even minor spatial deviations are captured, providing a level of layout-sensitivity that standard transformers lack.
\item \textbf{Green AI \& Environmental Sustainability:} Aligned with the Green AI initiative, CM$^2$ prioritizes computational efficiency. By avoiding the energy-intensive pre-training and massive carbon footprint associated with ``Red AI'' (large-scale LLMs), it offers an environmentally sustainable alternative for high-volume document processing.
\item \textbf{Optimized for CPU-Only Environments:} While modern NLP trends often rely on expensive, high-end GPU clusters, CM$^2$ is purpose-built to run on standard consumer-grade CPUs. Using static embeddings, such as GloVe, with $O(1)$ lookup latency rather than complex transformer passes, ensures near-zero inference time on basic hardware.
\item \textbf{Inherent Explainability (Glass-Box Model):} Unlike the ``black-box'' nature of deep learning and LLMs, CM$^2$ is fully interpretable. Every classification decision can be traced back to specific coordinate markers in the source document. This transparency is mandatory in audit-compliant industries such as finance, banking, and law.
\item \textbf{Faster Computation \& Low Latency:} The algorithm is designed for high-speed industrial throughput. By processing only ``important features'' rather than every word in a document, and by avoiding the heavy clock-cycle requirements of neural network layers, it provides a much faster classification pipeline than modern transformer benchmarks.
\item \textbf{Robustness Against Unbalanced Classes:} Because the model relies on structural identification rather than statistical word frequency, it is inherently robust against unbalanced datasets. It does not suffer from the ``majority class bias'' that often plagues traditional machine learning classifiers.
\item \textbf{Economic Viability:} By eliminating the need for expensive GPU infrastructure, high-cost LLM API subscriptions, and the labor-intensive process of large-scale data labeling, CM$^2$ provides a significantly cheaper solution for enterprise-level deployment.
\item \textbf{Generic, Expandable, and Extendable:} The coordinate matrix framework is highly versatile. It can be easily adapted to incorporate any structured document templates or extended to include additional feature markers without requiring the model to be ``re-trained'' on a massive corpus.
\end{enumerate}

\subsection{Organization}
The remainder of the paper is organized as follows. In Section~\ref{work}, we briefly discuss the related work in document classification. Our algorithm is formally described in Section~\ref{algo}. Section~\ref{exp} presents the design of the experimental investigation. We present and discuss the results in Section~\ref{result} and~\ref{discuss}. Finally, in Section~\ref{conc}, we suggest future improvements and conclude the paper.

\section{Related Work}\label{work}
As document classification has become an emerging area in text mining research, a large amount of prior work has focused on it. A typical document classification process has several pre-processing steps, and researchers have focused on each step to improve performance, such as stopword removal~\cite{manalu2017,prathibha2015}, Tokenization~\cite{bakar2014,gadri2015}, Part-of-Speech Tagging~\cite{owoputi2012}, and Stemming~\cite{zhang2007}.

The second step usually focuses on feature extraction and selection. For example, Yang et al.~\cite{yang2002} used titles and other tag data to label the text features. Shih and Karger~\cite{shih2004} used the geometry of the rendered HTML page to construct tree models based on the incoming-link structure. Term Frequency–Inverse Document Frequency (TF-IDF) is a widely used method for feature selection~\cite{jones1972}. Power et al.~\cite{power2010} focused on web page classification and filtered the output of the TF-IDF algorithm to improve the performance.

Once the feature vector is identified, the third step is to apply a machine learning model for classification. Na\"ive Bayes is one of the probability-based classifiers~\cite{kotenko2015}. Decision tree-based approaches have also been used for various purposes, such as blocking inappropriate web content~\cite{liu2017}. Support Vector Machines (SVM) are also widely used for document classification~\cite{farhoodi2010,lin2006}. There are multiple approaches under artificial/deep neural networks. Hassan and Mahmood~\cite{hassan2017} applied Convolution Neural Networks (CNN) for document classification. Unlike most machine learning models, Recurrent Neural Networks (RNNs) consider the sequence of word occurrences; therefore, documents with similar words can yield different outputs because of word order~\cite{banerjee2019,kumar2019}. The critical distinction in our methodology is that, whereas most approaches consider text and semantic proximity, we treat both text and coordinates as primary features. This enables CM$^{2}$ to excel in environments where the physical location of a token is as meaningful as the token itself.

While the landscape of document classification is currently dominated by Large Language Models (LLMs) and transformer-based architectures, these ``Red AI'' approaches~\cite{schwartz2020green} come with significant trade-offs in computational power, high-end GPU requirements, and interpretability. Recent research on Small Language Models (SLMs), such as DistilBERT~\cite{sanh2019distilbert}, and on efficient few-shot frameworks, such as SetFit~\cite{tunstall2022setfit}, aims to mitigate these costs by reducing model size; however, they remain fundamentally sequence-based. These models process text as a linear flow of tokens, often overlooking the rigid geometric structure and precise spatial coordinates that define formal documents such as bank statements or invoices.

A fundamental gap remains: standard transformers infer context from semantic proximity, whereas CM$^2$ directly leverages coordinate-based structural intelligence. By treating a document as a matrix of coordinates rather than a linear sequence, we provide a solution that is purpose-built for high-stakes industrial applications. This alignment with the Green AI philosophy ensures that the framework is not only hardware-agnostic and economically viable for CPU-only deployment, but also inherently interpretable~\cite{rudin2019stop}, addressing the strict transparency and structural requirements mandated in the financial and legal sectors.

\section{Methodology}\label{algo}
We now describe in more detail the steps of our algorithm for constructing the coordinate matrix and inducing the classifier from it.

\subsection{Pre-processing the Documents}
Each training data point is a PDF file containing either a scanned image or a digital document. We have ensured that each page of the document is set to 300 dpi before running it through an Optical Character Recognition (OCR) engine to obtain the words and their corresponding coordinates. The OCR output is stored as eXtensible Markup Language (XML) files, which are used as input to train our model.

\subsection{Building the Coordinate Matrix}
For each class, only one sample is required for training. We accompany each XML file with a Comma-Separated Value (CSV) file that contains a key-value pair for each keyword in the document.
\begin{example}
Statement A contains the term `Account No.’ followed by the account number `061234-12345678’, the term `Account Holder’ with the value `John Doe’, and the term `Account Type’ with the value `Savings Account’. Whereas, in statement B, the relevant term `Account No.’ followed by the number `064321-87654321’, and the term `Account Name’ with the value `Jane Smith’ is recorded. So in the CSV file for statement A, we have the entries ``\texttt{Account No.,061234-12345678}'', ``\texttt{Account Holder,John Doe}'' and ``\texttt{Account Type,Savings Account}''; and for statement B, we have ``\texttt{Account No.,064321-87654321}'' and ``\texttt{Account Name,Jane Smith}'' recorded. We then search for keywords (e.g., `Account No.', `Account Name') in the XML file and construct a matrix of top and left positions for each keyword in each document, as shown in Tab.~\ref{matrix}.
\end{example}
\begin{table}[t]
\centering
\caption{Example of the Coordinate Matrix}\label{matrix}
\begin{tabular}{llrr}
\toprule
Document ID & Keyword & Top & Left\\
\midrule
Statement A & Account No. & 254 & 1231\\
Statement A & Account Holder & 261 & 1231\\
Statement A & Account Type & 269 & 1231\\
Statement B & Account No. & 1123 & 231\\
Statement B & Account Name & 100 & 359\\
\bottomrule
\end{tabular}
\end{table}

The first column comes from the training data, the second column comes from the CSV file, and each XML file is accompanied by the third column, which contains the search results for the keywords in the XML files. Algorithm~\ref{ccm} lists the steps.
\IncMargin{1.5em}
\begin{algorithm}[b!]
\SetKwData{CM}{$\mathcal{CM}$}
\SetKwData{counter}{$row$}
\SetKwFunction{GetCoordinates}{GetCoordinates}
\SetKwInOut{Input}{Input}
\SetKwInOut{Output}{Output}
\SetKw{Return}{return}
\Input{$\mathcal{T}= \{t_1, t_2, \ldots, t_N\}$, where $t_i$ is a training sample of $i^{th}$ class and $N$ is the total number of class.\newline
$\mathcal{K}=\{K_1, K_2, \ldots, K_N\}$, where $K_i = \{k^i_1, k^i_2, \ldots k^i_{M_i}\}$ is a set of keywords for $i^{th}$ class, $k^i_j$ is the $j^{th}$ keyword of $i^{th}$ class, and $M_i$ is the total number of the keywords for $i^{th}$ class.}
\Output{\CM: Matrix of coordinates for all keywords for all training samples, i.e., $\langle t_i, k^i_j, x_{ij}, y_{ij} \rangle$ for every $1 \leqslant i \leqslant N$, $1 \leqslant j \leqslant M_i$}
\BlankLine
\Begin{
$M \leftarrow \sum_{i=1}^{N} M_i$\tcp*[r]{total number of keywords}
$\counter \leftarrow 1$\;
$\CM \leftarrow M \times 4$ matrix\;
\For{$i \leftarrow 1$ to $N$}{
	\For{$j \leftarrow 1$ to $M_i$}{
		$\langle x_{ij}, y_{ij} \rangle \leftarrow \GetCoordinates(t_i, k^i_j)$\;
		$\CM_{\counter} \leftarrow \langle t_i, k^i_j, x_{ij}, y_{ij} \rangle$\;
		$\counter \leftarrow \counter + 1$\;
	}
}
\Return $\CM$\;
}
\BlankLine
\BlankLine
\tcp{Function to get the coordinates of a keyword}
\Fn{\GetCoordinates{$\mathcal{D}$, $k$}}{
\Input{$\mathcal{D}$: a document as XML\newline
$k$: the keyword to find in the document.}
\Output{$\langle x, y \rangle$: the coordinates of the start position of the keyword $k$ in the document $\mathcal{D}$.}
\BlankLine
\If{$k \in \mathcal{D}$}{
\Return $\langle x, y \rangle$ \tcp*[r]{the keyword is found}
}
\Return $\varnothing$\;
}
\caption{Build the Coordinate Matrix\label{ccm}}
\end{algorithm}
\DecMargin{1.5em}

\subsection{Classifying New Documents}
When a new document arrives, we run OCR on it. In the XML produced by the OCR, we identify all words in the test document. We do this to ensure the approach remains robust to rotations or shifts in the coordinates that can occur during document scanning.

We match the extracted words against all the keywords in the training data. Once we extract the matched keywords, we then form a coordinate matrix for the test cases, comprising the top and left positions. 
\begin{example}
Test case contains the term `Account No.’ followed by the account number `061111-11111111’, and the term `Account Name’ with the value `Jane Doe’. Thus, the coordinate matrix for the test case is shown in Tab.~\ref{matrix2}. The second column of the table lists all keywords present in the training coordinate matrix.
\end{example}
\begin{table}[tb]
\centering
\caption{Coordinate Matrix for the Test Case}\label{matrix2}
\begin{tabular}{llrr}
\toprule
Document ID & Keyword & Top & Left\\
\midrule
Test case & Account No. & 1120 & 230\\
Test case & Account Holder & -- & --\\
Test case & Account Type & -- & --\\
Test case & Account Name & 101 & 360\\
\bottomrule
\end{tabular}
\end{table}

We then compute the distance between each keyword in the test data and every training sample. To calculate the distance between keywords, the Manhattan distance was used. We used Manhattan distance because it places greater weight on horizontal and vertical shifts than on diagonal shifts. The horizontal and vertical shifts are more likely in a document due to an extra empty line or space.

We have introduced only one parameter in this algorithm.
\begin{definition}
\textbf{Maximum Penalty} is the maximum distance allowed between two keywords. If the distance between the same keywords from two documents is more than this threshold, then the actual distance is substituted by this value. Besides, when a keyword is not found in a document, the distance for that keyword is set to this value. In other words, if the distance between the keyword found and its counterpart is more than the \textbf{maximum penalty}, we assume that the keyword is not found in the document.
\end{definition}

We defined this parameter for two reasons:
\begin{enumerate}
\item There should be a distance value even when we cannot locate a keyword in a document, otherwise we cannot apply the mean function in the next step.
\item We aim to ensure the algorithm’s robustness by limiting the impact of a single keyword. Otherwise, if a keyword in two documents is very far apart, either due to poor OCR extraction quality or because it is a variant of the training data, the calculated distance will be too large.
\end{enumerate}

Finally, we compute the mean distance for all keywords in each training sample to obtain a similarity score between the test case and that sample. The sample with the minimum distance to the test case is assigned to the class. Algorithm~\ref{classify} lists the steps, and Example~\ref{example} demonstrates the calculation.
\IncMargin{1.5em}
\begin{algorithm}[t!]
\SetKwData{DIJ}{$D_{ij}$}
\SetKwData{Class}{$\mathcal{C}$}
\SetKwData{Distance}{$\Delta$}
\SetKwData{MaxPen}{$\theta$}
\SetKwData{Doc}{$\mathcal{D}$}
\SetKwData{CM}{$\mathcal{CM}$}
\SetKwFunction{GetCoordinates}{GetCoordinates}
\SetKwInOut{Input}{Input}
\SetKwInOut{Output}{Output}
\SetKw{Return}{return}
\Input{\CM: the Coordinate Matrix which contains $\langle t_i, k^i_j, x_{ij}, y_{ij} \rangle$ for each rows\newline
\Doc: the test document as XML\newline
\MaxPen: maximum\_penalty, the maximum distance allowed between two keywords}
\Output{$\langle\Class, \Distance\rangle$, where \Class is the predicted class and\newline
\Distance is the minimum distance between the training and the test document}
\BlankLine
\Begin{
\For{$i \leftarrow 1$ to $N$}{
	\For{$j \leftarrow 1$ to $M_i$}{ 
		\tcp{\DIJ represents the distance from $k^i_j$ to its counterpart in \Doc}
		$coords \leftarrow \GetCoordinates(\Doc, k^i_j)$\;
		\eIf{$coords = \varnothing$}{
			$\DIJ \leftarrow \MaxPen$ \tcp*[r]{the keyword is not found}
		}
		{
			$\langle tx_{ij}, ty_{ij} \rangle \leftarrow coords$\;
			$\DIJ \leftarrow | x_{ij} - tx_{ij} |+| y_{ij} - ty_{ij} |$\;
			\If{$\DIJ > \MaxPen$}{
				$\DIJ \leftarrow \MaxPen$\;
			}
		}
	}
}
\tcp{Identifying the class}
$\Class \leftarrow \varnothing$\;
$\Distance \leftarrow \MaxPen$\;
\For{$i \leftarrow 1$ to $N$}{
	$distance \leftarrow \sum_{j=0}^{M_i} \DIJ\times\frac{1}{M_i}$\;
	\If{$distance < \Distance$}{
		$\Class \leftarrow t_i$\;
		$\Distance \leftarrow distance$\;
	}
}
\Return $\langle \Class, \Distance \rangle$\;
}
\caption{Classifying New Documents\label{classify}}
\end{algorithm}
\DecMargin{1.5em}

\begin{example}\label{example}
Let us assume that the \texttt{maximum\_penalty} is 200. Table~\ref{distance} shows the training data and the test case for each keyword.\\
\begin{table}[b]
\centering
\footnotesize
\caption{The Distance between the Training Data and the Test Case for Each Keyword}\label{distance}
\begin{tabular}{llrr}
\toprule
Document ID & Keyword & Calculation & Distance\\
\midrule
Statement A & Account No. & $| 254 - 1120 | + | 1231 - 230 |$ & 200\\
Statement A & Account Holder & Not found & 200\\
Statement A & Account Type & Not found & 200\\
Statement B & Account No. & $| 1123 - 1120 | + | 230 - 231 |$ & 4\\
Statement B & Account Name & $| 101 - 100 | +| 360 - 359 |$ & 2\\
\bottomrule
\end{tabular}
\end{table}

The distance between `Account No.' in Statement A is 200 because the Manhattan distance exceeds 200 pixels. For `Account Holder' and `Account type', the distances are 200 because these fields are not present in the test case. As a result, the distance between the test case and Statement A is 200, while this value is the average of 4 and 2 for Statement B. Therefore, the test case belongs to Statement B class with a similarity score of 3.
\end{example}

\subsection{Complexity of the Algorithm}
Let us consider a training dataset $\mathcal{T}$ of $N$ samples, where each sample comprises several keywords and $M$ is the total number of keywords in all $N$ samples. The time complexity to classify one test case containing $L$ number of words would be $O(LM)$, because for each of the $M$ available keywords, which can include several words, we have to search the test document. We can see that the computational complexity of our algorithm is linear to the size of the test data. Assuming that the number of keywords in each document is the same or within a range, $M$ is proportional to $N$ (i.e., $M \sim N$), and thus $O \sim N$. This means that the computational complexity of our algorithm is also linear in the number of training samples.

\section{Experiments}\label{exp}
In our experimental analysis, we compare six vectorization and nine classification techniques. To ensure our research remains grounded in Green AI principles---prioritizing environmental sustainability and hardware accessibility---we intentionally exclude high-parameter LLMs that require GPU-intensive inference. Instead, we evaluate purpose-built small models and efficient benchmarks.

Our selection includes three Bag of Words vectorizer techniques (Term Frequency Vectorizer~\cite{zhang2010}, TF-IDF Vectorizer~\cite{jones1972}, Hashing Vectorizer~\cite{hashing,scikit-learn}), two word embedding techniques (Global Vector (GloVe)~\cite{pennington2014}'s 6B pre-trained tokens, and Google’s pre-trained Word2Vec~\cite{mikolov2013}), and one paragraph embedding technique (Doc2Vec~\cite{le2014}). We specifically chose GloVe (Global Vectors for Word Representation) over modern transformer-based benchmarks like DistilBERT~\cite{sanh2019distilbert}. Although DistilBERT is a ``distilled'' and smaller model, it still requires multiple transformer-layer passes, which consume significant CPU clock cycles. In contrast, GloVe provides static embeddings that function as a high-speed lookup table ($O(1)$ inference complexity). This choice allows us to maintain near-zero latency and a minimal carbon footprint, ensuring that our results are directly applicable to resource-constrained, CPU-only environments common in on-premises industrial data processing, where infrastructure costs and processing speed are mandatory constraints.

The classification algorithms used are: Logistic Regression, Decision Tree (with CART and C4.5), SVMs (with linear, polynomial, RBF, and Gaussian kernel), Random Forest, Na\"ive Bayes, k-Nearest Neighbour. We also compared against several deep learning classifiers: an Artificial Neural Network, a CNN, and an LSTM-based RNN.

\subsection{Data Set}
\begin{figure}[H]
\centering
\psfrag{CBA}[r][r][0.55]{Commonwealth Bank}
\psfrag{WPC}[r][r][0.55]{Westpac Bank}
\psfrag{NAB NAB NAB NAB NAB NAB}[r][r][0.55]{National Australia Bank}
\psfrag{ING}[r][r][0.55]{ING Bank}
\psfrag{SCP}[r][r][0.55]{Suncorp Bank}
\psfrag{Bank}[r][r][0.75]{Bank}
\psfrag{Number of statements}[t][][0.75]{Number of Statements}
\psfrag{200}[][][0.55]{200}
\psfrag{150}[][][0.55]{150}
\psfrag{100}[][][0.55]{100}
\psfrag{50}[][][0.55]{50}
\psfrag{0}[][][0.55]{0}
\includegraphics[width=0.75\textwidth]{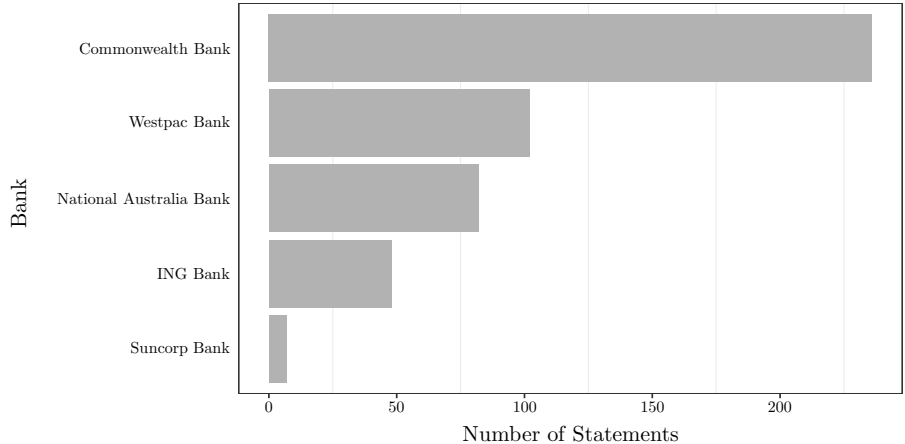}
\caption{The Distribution of Statements for Each Bank} \label{banks}
\end{figure}

We have randomly selected 475 statements from five banks that were crowdsourced for this research. The distribution of the banks’ statements is shown in Fig.~\ref{banks}. These statements fall under 53 templates. 16 of which have only 1-2 samples.

\subsection{Experiment Design}
Each of the vectorizers is applied with each classifier, except Na\"ive Bayes, which we could not use with Hashing, Google’s Word2Vec, and Doc2Vec vectorizers (because it is not compatible with them), making a total of 51 combinations to compare against our approach.

We have set aside 127 samples as test data while optimising the parameters using grid search. And used the remaining 348 to train models. We have ensured that for the minority classes, at least one sample remains in the training set. We employed 10-fold cross-validation (CV) for the machine learning models and 3-fold CV for the deep learning models to select the best parameters on the training set. Table~\ref{parameters} shows all the parameter values used during the grid search. Once the best parameters are identified, we train the final model on the entire training set and evaluate it on 127 test data points.
\begin{table}[t]
\centering
\footnotesize
\caption{Parameters Used during Grid Search for Each Algorithm from Sci-kit Learn~\cite{scikit-learn} and Keras~\cite{chollet2015}}\label{parameters}
\begin{tabular}{ll}
\toprule
Algorithm & Parameters\\
\midrule
Frequency Vectorizer, & \texttt{ngram\_range}: [1, 4)\\
TF-IDF Vectorizer, & \texttt{max\_features}: [20, 40, 60, \ldots, 6800]\\
Hashing Vectorizer & \\
\midrule
GloVe, & \texttt{max\_features}: [20, 40, 60, \ldots, 6800]\\
Google’s Word2Vec, & \\
Doc2Vec & \\
\midrule
Logistic Regression & \texttt{penalty}: \{none, $\ell_2$\}\\
& \texttt{C}: \{0.000001, 0.009, 0.001, 0.09, 0.01, 1, 5, 10, 25\}\\
& \texttt{max\_iter}: \{100, 120, 130, 140, 150\}\\
\midrule
Decision Tree & \texttt{criterion}: \{gini, entropy\}\\
& \texttt{min\_samples\_split}: \{2, 4, 5, 6\}\\
& \texttt{max\_features}: \{auto, sqrt, $\log$2, none\}\\
\midrule
Support Vector Machine & \texttt{C}: \{0.001, 0.01, 0.1, 1, 10, 100\}\\
& \texttt{kernel}: \{linear, poly, rbf, sigmoid\}\\
& \texttt{degree}: \{1, 2, 3\}\\
& \texttt{tol}: \{0.0001, 0.001, 0.1\}\\
& \texttt{decision\_function\_shape}: \{ovo, ovr\}\\ 
\midrule
Random Forest & \texttt{n\_estimators}: \{10, 11, 13, 15, 100, 115, 120, 125,\\
& 150, 200\}\\
& \texttt{criterion}: \{gini, entropy\}\\
& \texttt{min\_samples\_split}: \{2, 4, 5, 6\}\\
& \texttt{max\_features}: \{auto, sqrt, log2, None\}\\
\midrule
Na\"ive Bayes & \texttt{alpha}: \{0., 0.0001, 0.001, 0.01, 0.1, 1, 10\}\\
& \texttt{fit\_prior}: \{True, False\}\\
\midrule
$k$-Nearest Neighbour & \texttt{n\_neighbors}: \{3, 5, 7, 9, 11, 13\}\\
& \texttt{algorithm}: \{ball\_tree, kd\_tree, brute\}\\
& \texttt{leaf\_size}: \{30, 35, 30, 45, 50, 55\}\\
\midrule
Artificial Neural Network, & \texttt{activation}: \{relu, $\tanh$\}\\
CNN, & \texttt{optimizer}: \{SGD, Adam, Adamax, Adagrad,\\
LSTM-based RNN & Adadelta, Nadam, RMSprop\}\\
& \texttt{epochs}: \{10, 50, 100\}\\
& \texttt{learn\_rate}: \{0.01, 0.1, 0.2\}\\
\bottomrule
\end{tabular}
\end{table}

To assess the effect of training set size, we trained each model on 53 samples and evaluated its performance on the remaining data. We then augment the training data with 59 samples and evaluate the model’s performance on the remaining data. Eventually, we trained each model on 53, 112, 171, 230, 289, and 348 training data points and tested its performance on the remaining data. 

\section{Results}\label{result}
\begin{figure}[t]
\psfrag{CM2}[l][l][0.75]{CM$^2$}
\psfrag{TFVECT TFVECT TFVECT TFVECT}[l][l][0.6]{Term Frequency Vectorizer}
\psfrag{TFIDFVECT TFIDFVECT}[l][l][0.6]{TF$-$IDF Vectorizer}
\psfrag{HASHINGVECT HASHINGVECT}[l][l][0.6]{Hashing Vectorizer}
\psfrag{GLOETFIDFVECT GLOETFIDFVECT GL}[l][l][0.59]{GloVe with TF$-$IDF Vectorizer}
\psfrag{GOOGLEORD2VEC}[l][l][0.6]{Google’s Word2Vec}
\psfrag{DOC2VEC}[l][l][0.6]{Doc2Vec}
\psfrag{VECT}[r][][0.85]{Vectorizer}
\psfrag{FMEASURE}[][][0.85]{$F$-measure (\%)}
\psfrag{SIZE OF TRAINING}[t][][0.85]{Size of Training Data Set}
\psfrag{LR}[][][0.65]{Logistic Regression}
\psfrag{DT}[][][0.65]{Decision Tree}
\psfrag{SVM}[][][0.65]{Support Vector Machine}
\psfrag{RF}[][][0.65]{Random Forest}
\psfrag{NB}[][][0.65]{Na\"ive Bayes}
\psfrag{KNN}[][][0.65]{k-Nearest Neighbour}
\psfrag{ANN}[][][0.65]{Artificial Neural Network}
\psfrag{CNN}[][][0.65]{Convolutional Neural Network}
\psfrag{LSTM}[][][0.65]{LSTM-based Recurrent Neural Network}
\psfrag{348}[b][][0.63]{348}
\psfrag{289}[b][][0.63]{289}
\psfrag{230}[b][][0.63]{230}
\psfrag{171}[b][][0.63]{171}
\psfrag{112}[b][][0.63]{112}
\psfrag{53}[b][][0.63]{53}
\psfrag{0.00}[r][r][0.65]{0}
\psfrag{0.25}[r][r][0.65]{25}
\psfrag{0.50}[r][r][0.65]{50}
\psfrag{0.75}[r][r][0.65]{75}
\psfrag{1.00}[r][r][0.65]{100}
\includegraphics[width=\textwidth]{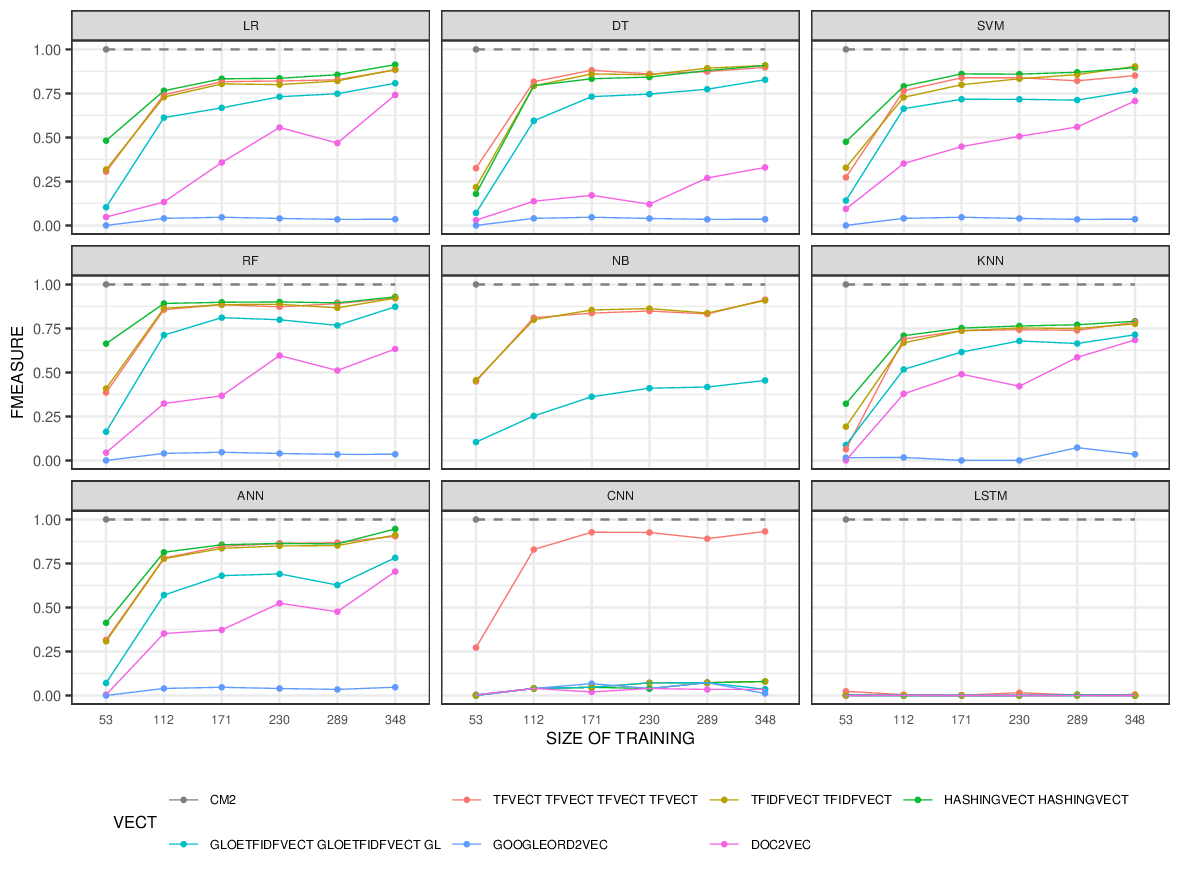}
\caption{Performance of Different Vectorizer with Every Classifier Compared Against CM$^2$ for Varying Training Data Set}\label{resultset}
\end{figure}

Figure~\ref{resultset} shows the results for different algorithms used with different vectorization methods while the size of the training data is growing. 
Each chart corresponds to a single machine learning approach, and the colours denote the vectorization method. As expected, the line charts show an upward trend, indicating that performance improves with increasing numbers of training samples. This upward trend also demonstrates the trade-off between labelled data and performance.

On average, the Hashing Vectorizer outperforms other vectorization methods for most classifiers, particularly with smaller training datasets. Google’s Word2Vec yields the worst performance across all selected classifiers. This is primarily because banks sometimes use non-standard abbreviations in transaction descriptions to fit longer words into a fixed-width text field, and Google’s pre-trained model is not designed for this.

The ensemble classifier Random Forest outperforms all other classification techniques, closely followed by a simple ANN. In contrast, LSTM performed very poorly because it requires large amounts of training data and long word sequences. In our experiment, the LSTM achieved training accuracy over 98\% on 348 documents, but it struggled to identify more than two samples correctly during testing.  Furthermore, bank statements do not contain any sentences that could benefit from techniques such as LSTMs.
 
For this experiment, we run our algorithm only once because we use only 53 samples, and the result is shown as a grey dashed line in the charts. As we can see, our method outperforms all machine learning methods, regardless of classifier and vectorizer type and training data size. Note that all other models are run with the best parameters obtained from the grid searches. Although we have reported the best performance of these models using the best parameters, our algorithm still yields better performance with a simple setting and the smallest training data set. This shows that selecting appropriate features and designing an appropriate approach are more important than switching or ensembling models.

\begin{figure}
\centering
\psfrag{MAX}[t][][1]{Maximum Penalty}
\psfrag{FMEASURE}[b][][1]{$F$-measure (\%)}
\psfrag{500}[][][0.65]{500}
\psfrag{400}[][][0.65]{400}
\psfrag{300}[][][0.65]{300}
\psfrag{200}[][][0.65]{200}
\psfrag{100}[][][0.65]{100}
\psfrag{90}[r][r][0.65]{90}
\psfrag{92}[r][r][0.65]{92}
\psfrag{94}[r][r][0.65]{94}
\psfrag{96}[r][r][0.65]{96}
\psfrag{98}[r][r][0.65]{98}
\psfrag{10}[r][r][0.65]{100}
\includegraphics[width=0.75\textwidth]{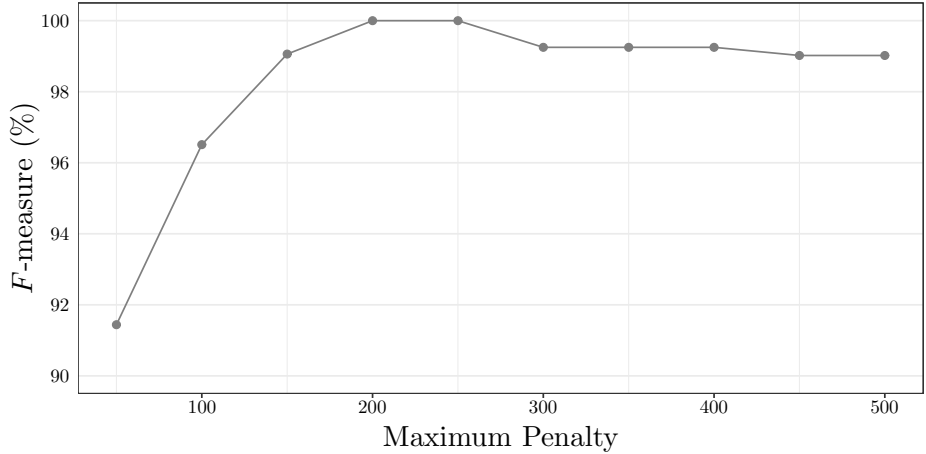}
\caption{Effect of Different Maximum Penalty on CM$^2$ Performance}\label{penalty}
\end{figure}

Figure~\ref{penalty} shows the parameter sensitivity analysis for the maximum penalty, which is the only parameter required by our algorithm.
We observe that, at the minimum, performance is lower because our approach is sensitive to the coordinates of the keywords. In other words, if the keyword is just a little shifted from its expected position, we assume the keyword is not found. In practice, the locations of keywords may change due to scanning or the presence of extra lines or spaces. At \texttt{maximum\_penalty} = 200, the accuracy is 100\%. Let us consider the resolution of an input image 2480 $\times$ 3500. If the keyword is around 10\% of the width apart from the original location of the keyword, it is considered not found.

By increasing the maximum penalty by more than 250, we also observe a drop in accuracy because keywords that are relatively far apart are now considered matched. After 400, we do not observe a significant difference in performance. The reason is that when we set a very high threshold, only a few distances fall below it. Theoretically, all the Manhattan distances in a 2480 $\times$ 3500 document are less than 2480 + 3500. Therefore, setting a threshold above 2480 + 3500 is unnecessary and has no effect.

\section{Discussion \& Sustainability}\label{discuss}
We have demonstrated that, when appropriately designed, human-level concept learning is far superior to machine learning methods, especially when learning from limited data. In contrast to machine learning methods, which require hundreds of samples per class to learn the concept, our approach requires only one sample and leverages the concept more richly. As with humans, our approach attempts to match the test case to each known sample, estimate similarities, and select the most similar class. Upon closer inspection, our approach identifies keywords and matches them across documents. However, if the matched keywords are too far apart, the match is discarded, and the keyword is assumed to have no match. Similarly, as humans, we would not match keywords that are too far apart to detect a document type.

By avoiding the ``black-box'' nature of large-scale transformer architectures, CM$^2$ provides inherent interpretability and a significantly lower carbon footprint. This aligns our work with modern requirements for audit-compliant and sustainable AI, proving that specialized structural intelligence can outperform general-purpose models in niche industrial domains.
 
Another advantage of the human-level concept learning is the way we set the parameters. In human-level concept learning, our understanding of the problem helps us define the values of the algorithm parameters. In contrast, in machine learning approaches, a parameter optimization technique (e.g., grid search) is required to find the best values. This can be very time-consuming and depends on the number of parameters to tune, the number of values to consider for each parameter, or the algorithm’s computational complexity. Furthermore, the algorithm must be run for each combination of parameter settings. Still, there is no guarantee that the best parameter value is among the considered parameter values.

When we want to set a value for \textit{maximum penalty}, we should ask ourselves this question ``what is the maximum distance that we want to allow to match keywords?'' or How far a keyword can shift due to some extra line or extra space in a document? Based on sample documents, we estimate that $200$ is an appropriate value for the \textit{maximum penalty}. Definitely, it is better to check the other values, but we expect the best value to be around $200$. On the other hand, assume we are using a deep learning model and want to set the values of $\ell_1$ and $\ell_2$. It is difficult to interpret these values in the context of our problem. Therefore, we must choose them using trial-and-error or a grid search. In most cases, grid search with more parameters tends to overfit~\cite{cawley2010}.
 
Designing human-level concept learning begins by considering how humans perform the task. First, we examine the data, and then we should ask ourselves, as humans, how we solve the challenge. How do we recognize a template for bank statements whose contents are almost identical? The answer to the question identifies which feature to use and how to design the approach.

\section{Conclusion}\label{conc}
In this paper, we introduced the Coordinate Matrix Machine (CM$^2$), a novel framework for one-shot document classification that mimics human-level concept learning. By subconsciously identifying important structural features rather than processing exhaustive semantic data, CM$^2$ successfully addresses the challenge of classifying highly similar documents with only a single sample per class.

Our experimental results demonstrate that CM$^2$ outperforms traditional vectorization techniques and complex deep learning models in accuracy, speed, and robustness. Beyond performance metrics, this research makes a significant contribution to the Green AI initiative. We have demonstrated that purpose-built small models can achieve superior results in specialized domains without the immense carbon footprint, energy consumption, and hardware costs associated with Large Language Models (LLMs).

Furthermore, CM$^2$ addresses the ``black-box'' limitations of modern neural network-based architectures. By prioritizing coordinate-based structural intelligence over linear sequences, our model provides inherent explainability, a mandatory requirement for audit-compliant industrial applications. We intentionally used static embeddings, such as GloVe, to maintain near-zero inference latency on standard CPUs, demonstrating that hardware-agnostic, low-cost AI is a viable and effective alternative to resource-heavy infrastructure.

Future work will explore the application of CM$^2$ in broader multi-modal contexts and its integration into real-time, high-throughput financial pipelines, continuing our commitment to sustainable, transparent, and efficient machine learning.

\section*{Acknowledgement}
The authors thank the AI4Convery team for crowdsourcing 500 bank statements, anonymising them by removing personally identifiable information (PII), providing the statements in PDF format, and running OCR to produce XML for this research.


\bibliographystyle{unsrt}
\bibliography{cm2}

\end{document}